\documentclass[runningheads]{llncs}
\usepackage[T1]{fontenc}
\usepackage{graphicx}
\usepackage{multirow}
\usepackage{xcolor}
\usepackage{tikz}
\usepackage{placeins}
\usepackage[utf8]{inputenc}
\usepackage{amsmath}
\usepackage{todonotes}
\usepackage{algpseudocode}
\usepackage{algorithm}
\usepackage{caption}
\usepackage{framed}
\usepackage{float}
\usepackage{amssymb}
\usepackage{amsfonts}
\usepackage{hyperref}
\usepackage{soul}
\usepackage{tikz}
\usepackage{comment}
\usepackage{graphicx}
\usepackage{subcaption}

\usepackage{color}

\newcommand{\SSS}{\mathcal{S}} 
\newcommand{\AAA}{\mathcal{A}} 
\newcommand{\RRR}{\mathcal{R}} 
\newcommand{\R}{\ensuremath{\mathbb{R}}}

\begin{document}

\title{A Systematization of the Wagner Framework: Graph Theory Conjectures and Reinforcement Learning}

\titlerunning{A Systematization of the Wagner Framework}

\author{Flora Angileri\inst{1}\orcidID{0009-0001-3968-6973} \and
Giulia Lombardi\inst{2, 9}\orcidID{0000-0002-6953-5447} \and
Andrea Fois\inst{3}\orcidID{0000-0002-2749-240X} \and
Renato Faraone\inst{3}\orcidID{0000-0003-2426-0299} \and
Carlo Metta\inst{4, 7}\orcidID{0000-0002-9325-8232} \and
Michele Salvi\inst{1, 8, 9}\orcidID{0000-0001-8519-4665} \and
Luigi Amedeo Bianchi\inst{2, 9}\orcidID{0000-0001-7040-0366} \and
Marco Fantozzi\inst{3}\orcidID{0000-0002-0708-5495} \and
Silvia Giulia Galfrè\inst{5}\orcidID{0000-0002-2770-0344} \and
Daniele Pavesi\inst{3}\orcidID{0009-0005-3073-5379} \and
Maurizio Parton\inst{6, 9}\orcidID{0000-0003-4905-3544} \and
Francesco Morandin\inst{3, 9}\orcidID{0000-0002-2022-2300}}

\authorrunning{F. Angileri, G. Lombardi, A. Fois, R. Faraone, C. Metta, M. Salvi et al.}

\institute{Tor Vergata University of Rome, Italy \and
University of Trento, Italy \and
University of Parma, Italy \and
ISTI-CNR, Pisa, Italy \and
University of Pisa, Italy \and
University of Chieti-Pescara, Italy}


\maketitle

\footnotetext[7]{EU Horizon 2020: G.A. 871042 SoBig-Data++, NextGenEU - PNRR-PEAI (M4C2, investment 1.3) FAIR and “SoBigData.it”.}
\footnotetext[8]{PRIN project Grafia (CUP: E53D23005530006), Department of Excellence MatMod@Tov (CUP: E83C23000330006).}
\footnotetext[9]{Funded by INdAM groups GNAMPA and GNSAGA.}

\let\oldthefootnote=\thefootnote
\renewcommand{\thefootnote}{}
\footnotetext{$^\text{all}$Computational resources provided by CLAI laboratory, Chieti-Pescara, Italy.}
\footnotetext{$^\text{all}$Authors can be contacted at \texttt{curiosailab@gmail.com}.}
\let\thefootnote=\oldthefootnote

\vspace{-2.5em}
\begin{abstract}
In 2021, Adam Zsolt Wagner proposed an approach to disprove conjectures in graph theory using Reinforcement Learning (RL). Wagner frames a conjecture as \( f(G) < 0 \) for every graph \( G \), for a certain invariant $f$; one can then play a single-player graph-building game, where at each turn the player decides whether to add an edge or not. The game ends when all edges have been considered, resulting in a certain graph \( G_T \), and \( f(G_T) \) is the final score of the game; RL is then used to maximize this score. This brilliant idea is as simple as innovative, and it lends itself to systematic generalization. Several different single-player graph-building games can be employed, along with various RL algorithms. Moreover, RL maximizes the cumulative reward, allowing for step-by-step rewards instead of a single final score, provided the final cumulative reward represents the quantity of interest \( f(G_T) \). In this paper, we discuss these and various other choices that can be significant in Wagner's framework.
As a contribution to this systematization, we present four distinct single-player graph-building games. Each game employs both a step-by-step reward system and a single final score. We also propose a principled approach to select the most suitable neural network architecture for any given conjecture and introduce a new dataset of graphs labeled with their Laplacian spectra.
The games have been implemented as environments in the Gymnasium framework, and along with the dataset and a simple interface to play with the environments, are available at \url{https://github.com/CuriosAI/graph_conjectures}.

\end{abstract}

\section{Introduction}

The field of graph theory is a wellspring of conjectures that have long fueled mathematical investigation.
Recently, Wagner in~\cite{WagCCN} proposed an innovative approach to disprove these conjectures, formulating the problem as a one-player game modeled within the Reinforcement Learning (RL) framework. In this game, the player maneuvers through a state space of graphs, earning rewards based on certain graph characteristics, and related to the conjecture in question. Through optimal play, the game steers the player towards a graph that is as close as possible to the conjectured bound. Surpassing the bound provides a counterexample to the conjecture.

In this very general framework, once a target conjecture is chosen, e.g. \( f(G) < 0 \) for every graph \( G \), there are several pivotal choices that could lead to success or failure. For instance, the rules of the ``build your graph'' game; the reward function; the termination condition; the RL algorithm used to play the game and optimize the cumulated reward; the neural network architecture involved in the RL algorithm. Each of these variables, and many others as well, has an impact on the model's capability to explore successfully the space of graphs and eventually finding a counterexample. Moreover, if the bound is not surpassed despite the player learning, something can still be inferred: the conjecture is true, and experiments gives us empirical evidence in favor of this, or the counterexample is rare with respect to the visitation distribution of the RL algorithm in use, and this suggests changing some of the choices toward a more sophisticated exploration.

\textbf{Novel Contributions.} The aim of this paper is to open a discussion about those pivotal choices: among various ``build your graph'' games, reward functions, RL algorithms, neural network architectures, are some better than others? We argue that the first and most effective choice is the game, and we provide open-source Gymnasium \cite{Gymn} implementations of four different graph-building games, that we call Linear, Local, Global, and Flip. An externally defined reward function makes them independent from the conjecture. Moreover, we argue that the second important choice is the neural network architecture, that should excel at extracting features informative for computing $f$. We recommend preliminary testing of various architectures on a supervised task related to $f$, selecting the one that performs best. We introduce a novel dataset of graphs labeled with their Laplacian spectra, which is particularly useful for conjectures related to eigenvalues. Furthermore, we present a novel counterexample for Conjecture 2.1 in \cite{WagCCN}.
With this contribution, we hope to steer the research toward a general systematization of Wagner's framework.

In Section~\ref{sec:related}, we briefly review the body of literature that has stemmed from Wagner's idea. In Section~\ref{sec:methods}, we discuss the most relevant choices that can be done. In Section~\ref{sec:envs}, we describe four ``build your graph'' games, implemented as Gymnasium environments: Linear (similar to Wagner's original game), Local, Global, and Flip (similar to the game described in~\cite{MAKFIL}). In Section~\ref{sec:dataset}, we describe the Laplacian spectra dataset. Finally, in Section~\ref{sec:future}, we discuss possible future developments.

\section{Related Work}
\label{sec:related}

The paper where this framework was first proposed by Wagner is~\cite{WagCCN}. Here the game is played on graphs with a fixed number of nodes $n$, and the $\frac{n(n-1)}{2}$ edges are enumerated in a predefined order. The agent starts from the empty graph $G_0$, and at turn $t$ decides whether or not to add edge $t$, building graph $G_t$. The agent does not receive any reward until the last turn $t=T=\frac{n(n-1)}{2}$, when the game ends and the agent receives a reward $f(G_T)$. Note that since the agent needs to know which edge to add at every turn, states contain both the graph $G_t$, and the turn $t$ as well (common in the RL finite horizon setting). The policy is modeled as a fully connected 3-layers neural network, and the RL algorithm used is the gradient-free cross-entropy method. Using this beautiful idea, Wagner was able to find counterexamples for several published conjectures, including a 19-nodes counterexample for a conjecture on the sum of the matching number and the spectral radius~\cite{ACHAuS,SteRAC}. In our paper, we provide a Gymnasium implementation of Wagner's game, that we call Linear, and a 18-nodes counterexample.

Wagner's approach has led to several important follow-up studies, including~\cite{MAKFIL}, co-authored by Wagner himself. This study tackles an extremal graph theory problem originally proposed by Erdős in 1975. Their focus is on identifying graphs of a specific size that maximize the number of edges while excluding 3- or 4-cycles. Utilizing AlphaZero\cite{silver2017alphazero}, they bootstrap the search process for larger graphs using optimal solutions derived from smaller ones, enhancing lower bounds across various sizes. Key innovations of their work include a new game, that they call edge-flipping game, and a novel Graph Neural Network architecture, called pairformer, which has proven particularly effective for this problem. However, they did not make the code for the game, the pairformer, or the AlphaZero configuration used in their experiments publicly available. In our paper, we provide Flip, a Gymnasium implementation of their edge-flipping game.

Another very interesting paper is~\cite{GAKRL1}, in which the authors reevaluate Wagner's method with an emphasis on enhancing its speed and stability. They reimplement from scratch Wagner's code, improving the readability, stability, and speed. They also successfully construct counterexamples for various conjectured bounds on the Laplacian spectral radius of graphs. Like our work, they implement an external reward function. Yet, the most important contribution of their paper is, in our opinion, the special attention given to computational performance. They observe that, since RL must process invariants for hundreds of thousands of graphs to achieve adequate convergence in learning, using NetworkX \cite{NetworkX} and/or numpy is suboptimal. They show that invoking Java code directly from Python significantly accelerates the invariants computation. Our experiments confirm that using networkX is quite slow, and using this approach is the most natural future development of our paper, see Section~\ref{sec:future}.

In ~\cite{GAKRL2}, authored by the same team as ~\cite{GAKRL1}, new lower bounds are established for several small Ramsey numbers. They continue to use Wagner's original framework, but with a slight modification to the RL algorithm. This aligns with our proposal to diversify Wagner's framework from multiple directions to effectively tackle various conjectures.

\section{Methods}
\label{sec:methods}

\subsection{Notation}

In RL a game is typically modeled as a Markov Decision Process (MDP), that is, a 4-tuple $(\SSS, \AAA, \RRR, p)$ consisting of state space, action space, rewards, and transition model $p:\SSS\times\AAA\to\Delta(\SSS\times\RRR)$, where $\Delta$ denotes the space of probability distributions. A transition $p(s',r|s,a)$ is the probability of reaching next state $s'$ with a reward $r\in\RRR\subset\R$ when the action $a$ is executed in the state $s$. An agent interacts with the environment by sampling actions from a policy $\pi:\SSS\to\Delta(\AAA)$. This agent-environment interaction gives rise to a sequence $S_0,A_0,R_1,S_1,A_1,R_2,\dots$, called \emph{trajectory}. Here $S_t, A_t, R_{t+1}, S_{t+1}$ denote the state at time $t$, the action executed at time $t$ sampled from $\pi(\cdot|S_t)$, the reward received at time $t+1$, and the state reached at time $t+1$, respectively. Moreover, if there are absorbing states reachable from every state under a uniform policy, the game is called \emph{episodic} and will eventually terminate; otherwise, it is called \emph{continuing} and the trajectory will never end, unless an additional termination condition is given. Sometimes, the game ends always at a fixed time $T$, and in this case it is called a \emph{finite time horizon} MDP. Our game features a customizable finite time horizon, which defaults to the number of edges. For this and other details, see Sections~\ref{sec:states}, \ref{sec:rewards}, \ref{sec:algorithms}, \ref{sec:architectures}, and \ref{sec:envs}.

Gymnasium~\cite{Gymn} is a maintained fork of OpenAI’s Gym~\cite{Gym} library, a popular open-source framework developed by OpenAI that provides a standardized set of games, called \emph{environments} in Gym, for testing and developing RL algorithms. The Gym library is designed to help researchers and developers to easily experiment with different RL algorithms and compare their performance across a wide variety of tasks. For this reason, we have chosen Gymnasium to implement the graph-building games in this paper.

Given a family $\mathcal{G}$ of graphs, we always assume the conjecture in a normal form:
\begin{equation}\label{eq:conjecture}
f(G) \leq 0 \text{ or } f(G) < 0 \qquad \forall G \in \mathcal{G}.
\end{equation}
In our environments, $\mathcal{G}$ is the family of all undirected unweighted graphs with a fixed amount $n$ of nodes, without multiple edges. Self-loops are optional.

\subsection{States, Actions, and Transitions}
\label{sec:states}

In Wagner's framework, the MDP is a graph-building game, and thus, the state $S_t$ always contains at least the current graph $G_t$. Sometimes, prior knowledge on the problem can suggest to restrict to certain graphs, for instance when one can prove that a counterexample, if it exists, must happen on trees. In this case, one could consider to restrict the family $\mathcal{G}$ of graphs visited during episodes. We designed our environments for general undirected graphs with a fixed amount $n$ of nodes, without multiple edges, and we added the option to include or exclude self-loops. See also Section~\ref{sec:future} for possible improvements of our environments that could allow to change the family $\mathcal{G}$.

In Wagner's framework, the agent visits one edge a time, under a predefined order. However, in general, there are several different ways in which the agent can move on the graph and select the edge (or the edges) to modify. For instance, some games can be single-action based, like the edge-flipping game in~\cite{MAKFIL} and our Flip game; other games can be multi-action, with both possibilities to modify an edge or not, like Wagner's game, and our Linear, Local, and Global games. Moreover, the agent can modify an edge in several ways: by adding or removing it, but also by leaving it as it is or changing it, or by flipping it.
A significant aspect of how actions are defined is their impact on the game's dynamics. In some cases, the game becomes \emph{monotonic}, meaning that each graph $G_{t}$ is a subgraph of $G_{t+1}$. For details on our implementation and how we take these variations into account, see~Section~\ref{sec:envs}.

Another important consideration in Wagner's framework is the transition model used. In RL, the environment might respond to actions in a stochastic manner, with transitions modeled by the conditional probability distribution $p(s', r|s, a)$. Incorporating a stochastic element in the graph-building game, where an action $A_t$ performed on graph $G_t$ could lead to various possible next states $G_{t+1}$, is an intriguing possibility, because this stochastic approach could enhance exploration within the model.
However, in our initial paper on systematizing Wagner's framework, we implemented only deterministic games, where applying the same action to a given graph consistently results in the same next graph.

\subsection{Reward}
\label{sec:rewards}

Given a conjecture as in \eqref{eq:conjecture}, a natural choice for the reward in the episodic setting is $r(G_T) = f(G_T)$ at the end $T$ of the episode, and $r(G_T) = 0$ elsewhere. We call this reward \emph{sparse}. Then, a counterexample is found when $r(G_T) > 0$ happens. A different choice is what we call the \emph{incremental} reward: at each time $t>0$, we receive the increment $f(G_t)-f(G_{t-1})$. If we start from a ``virtual'' graph $G_0$ with $f(G_0)=0$, then the cumulated incremental reward is exactly $f(G_T)$, as with the sparse reward. In certain cases, for instance with temporal difference algorithms, or when we are interested in understanding how a single action impacts on the graph, an incremental reward could prove useful.

Observe that in a graph-building game, reaching the end of the episode before checking for the counterexample is not efficient. Since the game is just a way to guide agent's search for a counterexample, one could perform the check after every action. In this case, again, a sparse reward seems less reasonable. We provide the option of performing this check after every action in our environments. Note that in particular when this check is enabled, performing the invariant computation with an external Java code as suggested in~\cite{GAKRL1}, is particularly useful. See Section~\ref{sec:future}.

Furthermore, note that alternative rewards and a discounted setting could also be considered. As long as the objective of maximizing cumulative rewards potentially results in a counterexample, these alternatives remain viable. Additionally, incorporating a discount factor offers the advantage of applicability in continuing formulations of the game, when a termination condition is not desired.However, designing a reward system in a discounted setting such that maximizing it reliably leads to a counterexample presents significant challenges.

\subsection{RL algorithm}
\label{sec:algorithms}

\begin{figure}[t]
    \centering
    \captionsetup{justification=centering}
    \includegraphics[width=6cm, height=5cm]{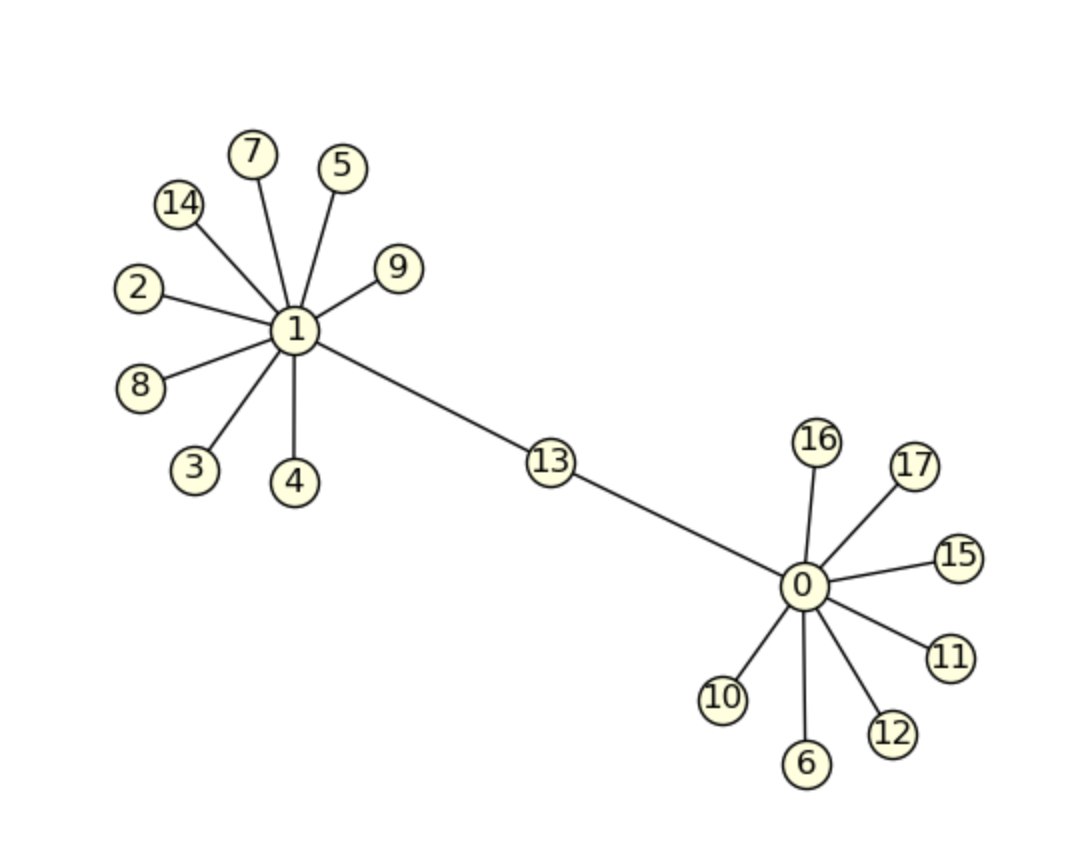}
    \caption{Counterexample $G$ for Conjecture 2.1 found with PPO. Here $f(G) = \sqrt{18-1} - 1 - \lambda_1(G) - \mu(G) \simeq 0.02181 > 0 $}
    \label{fig:counterex}
\end{figure}

The choice of the reinforcement learning (RL) algorithm is crucial. Wagner, and~\cite{GAKRL1,GAKRL2} also showed that even a simple, gradient-free algorithm like the cross-entropy method can achieve impressive results. Nonetheless, several alternative algorithms could also be considered. Among the most promising are Proximal Policy Optimization (PPO), known for its robustness and effectiveness across various tasks, and AlphaZero-like algorithms.
Employing PPO with our environments enabled us to find a novel 18-node counterexample to Conjecture 2.1 in Wagner's paper, see Fig.~\ref{fig:counterex}. Although this counterexample merely replicates the structure of Wagner's original, it underscores the potential applicability of different algorithms.

Despite PPO being promising, we posit that AlphaZero is the most natural choice in this highly complex graph-building game scenario, and in fact it was used successfully in~\cite{MAKFIL}.
However, notice that while PPO is available in several established RL library like Stable-Baselines \cite{stable-baselines3}, and its design space is relatively easy to configure, AlphaZero complexity requires a much more thorough calibration of its hyperparameters.

\subsection{Architecture of approximators}
\label{sec:architectures}

Given the combinatorial explosion of non-isomorphic graphs when increasing the number of nodes, approximation must be used. We assume that approximation is done by neural networks, and this takes the neural network architecture in play. Given the nature of the task, a clear ``best choice'' here is using a Graph Neural Network (GNN). But which GNN is the best one for a given conjecture $f$?

To guide the choice of a proper GNN architecture for a given conjecture $f$, we propose to test different architectures on a supervised task involving the most relevant invariants used in $f$. For instance, if the conjecture uses the largest eigenvalue and the matching number, one could train several GNN to predict the largest eigenvalue or the matching number of a graph, and then use in the graph-building game the GNN that in the supervised learning task provided the best accuracy. A dataset built to guide this selection process should be as rich as possible, including a lot of non-isomorphic graphs.

Following on this idea, we created a dataset of graphs with $11$ nodes, labelled with their Laplacian spectra, see Section~\ref{sec:dataset} for more details. This dataset is designed for Brouwer's Conjecture, that has been proven true up to $n=10$ in ~\cite{Brouw10}, see Section~\ref{sec:future}.

\section{Environments}
\label{sec:envs}

We implemented four ``build your graph'' games as Gymnasium environments: Linear, Local, Global, and Flip. All our games are played on undirected graphs with a fixed number $n$ of nodes, and without multiple edges. The environments are parametric with respect to several aspects: for instance, one can choose the starting graph, or whether to enforce the agent excluding self-loops. For details on these parameters, see Section~\ref{sec:common}. These environments are available at \url{https://github.com/CuriosAI/graph_conjectures}.

\subsection{Linear}

Linear is a variation of the game used by Wagner. The name comes from the state's vector internal representation. In Linear, edges are ordered, and then at each time $t$ the agent can choose between leaving the edge number $t$ as it is (i.e. passing it), or flipping it. The Edge-flipping operation (as defined in \cite{MAKFIL}) changes the state of an edge like a boolean \emph{not} operator, as follows: let $e \in \{0,1\}$ be the single bit representing the edge, then
\begin{align*}
    flip(e) = \begin{cases}
    1 \quad & \text{if } e=0 \\
    0 & \text{otherwise}
    \end{cases}
\end{align*}

The state is given by the graph and the current time $t$, and the action space is $\{0,1\}$, where $0$ means that the current edge is left unchanged, and $1$ that the edge is flipped. With its default values, Linear differs from Wagner's game for the ordering of the edges: in Wagner's game, edges are numbered by forming and expanding cliques first, that is, $(1,2), (1,3), (2,3), (1,4), \dots$, while in Linear is given by $(1,2), (1,3), \dots, (1,n), (2,3), \dots$.
Moreover, Wagner starts from the empty graph, while the default setting in our games is to start from the complete graph. Episodes in Linear always end at time $T=\frac{n(n-1)}{2}$, if self-loops are not allowed, and at time $T=\frac{n(n+1)}{2}$, otherwise.

\begin{figure}[htbp]
    \centering
    \begin{tikzpicture}[node distance={130mm},main/.style={circle,draw,color=blue!70,fill=blue!70,minimum size=2.5mm}]

\node[label=left:G] (0) at (-0.5,0.75) {};
\node[main, inner sep=1pt, label=left:1] (A) at (0,1.5) {};
\node[main, inner sep=1pt, label=right:2] (B) at (1.5,1.5) {};
\node[main, inner sep=1pt, label=right:3] (C) at (1.5,0) {};
\node[main, inner sep=1pt, label=left:4] (D) at (0,0) {};

\draw (A) -- (B);
\draw[green] (A) -- (D);
\draw (A) -- (C);
\draw[dotted,red,thick] (B) -- (C);
\draw[dotted] (D) -- (C);
\draw[dotted] (D) -- (B);

\node[main, inner sep=1pt, label=left:1] (A2) at (4.5,1.5) {};
\node[main, inner sep=1pt, label=right:2] (B2) at (6,1.5) {};
\node[main, inner sep=1pt, label=right:3] (C2) at (6,0) {};
\node[main, inner sep=1pt, label=left:4] (D2) at (4.5,0) {};

\draw (A2) -- (D2);
\draw[red] (B2) -- (C2);
\draw (A2) -- (C2);
\draw (A2) -- (B2);
\draw[dotted] (D2) -- (C2);
\draw[dotted] (B2) -- (D2);

\node[main, inner sep=1pt, label=left:1] (A3) at (4.5,-1.5) {};
\node[main, inner sep=1pt, label=right:2] (B3) at (6,-1.5) {};
\node[main, inner sep=1pt, label=left:4] (C3) at (4.5,-3) {};
\node[main, inner sep=1pt, label=right:3] (D3) at (6,-3) {};

\draw (A3) -- (B3);
\draw (A3) -- (D3);
\draw[dotted,green,thick] (A3) -- (C3);
\draw[dotted] (B3) -- (C3);
\draw[dotted] (B3) -- (D3);
\draw[dotted] (C3) -- (D3);

\draw[->] (2,0.75) -- (4,0.75);
\draw[->] (2,-0.25) -- (4,-1.25); 

\end{tikzpicture}
    \caption{\textit{Effects of the flip action on different edges}. Existing and missing edges are represented with solid and dotted lines, respectively.  The top-right graph shows \emph{G} after flipping edge (2,3), while the bottom-right graph shows \emph{G} after flipping edge (1,4).}
    \label{fig:action-flip}
\end{figure}

\subsection{Local}

In Local, the agent explores the graph space by moving from one node to another. When moving from node $i$ to node $j$, the agent has the option either to flip the edge $(i, j)$ or to pass it. This ensures that from node $i$, the agent's actions are ``locally'' confined, impacting only the directly connected edge $(i, j)$. Note that this is different from Linear, because the agent can choose any node $j$ to move to.
The state is given by the current graph, the current node $i$ where the agent is located, and the current time. An action is given by a target node $j$ to move to from node $i$, and a binary value $\{0, 1\}$, where $0$ means taking no action, and $1$ means flipping the edge $(i, j)$. In our implementation, this action logic is represented by a single integer value $k$ within the range $[0, 2n-1]$ where $n$ is the number of nodes. Assuming to start from node $i$, if $k \in [0, n-1]$, we move to node $j = k$ without taking any action on edge $(i,j)$. If $k \in [n, 2n-1]$, we move to node $j = k \mod n$, and the edge $(i,j)$ is flipped. Episodes in Local end at a termination time $T$ that can be passed as optional input when the game is initialized, and defaults to $T=\frac{n(n-1)}{2}$, if self-loops are not allowed, and to $T=\frac{n(n+1)}{2}$, otherwise.

\subsection{Global}

In Global, the agent explores the graph space by acting on any edge across the entire graph at any time. The agent can choose any edge to act upon, deciding either to flip it or to pass it. This ``global'' approach ensures that the agent's actions are not confined to its immediate location, allowing interaction with any part of the graph.
Episodes end at a termination time $T$ that can be given as input at game's initialization, with same Local defaults. Similar to Local, the possibility to pass on an action without flipping an edge is maintained, because it helps mitigate the risk of choosing a wrong termination time for the game. For instance, if flipping were mandatory, excessively long matches could potentially disrupt an optimal configuration previously achieved. Allowing the passing of actions enables the agent to maintain an optimal configuration indefinitely.
The state is given by the current graph, and the current time. An action is given by a target edge, and a binary value $\{0, 1\}$, where $0$ means taking no action, and $1$ means flipping the edge. In our implementation, the action logic is similar to that seen in Local, but generalized to handle global movements along the graph. Here, the action is represented by a single integer value $k$ within the range $[0, 2m-1]$, where $m$ is the number of edges. If $k \in [0, m-1]$, edge $(i, j)$,  where $i = \left\lfloor \frac{k}{n} \right\rfloor$ and $j = k \mod m$, remains unchanged. If $k \in [m, 2m-1]$, edge $(i, j)$, where $i = \left\lfloor \frac{k-m}{n} \right\rfloor$ and $j = k \mod m$ is flipped.

\subsection{Flip}

Flip implements the edge-flipping environment as described in~\cite{MAKFIL}. This environment is the same as Global, with one difference: the absence of the option to pass. In Flip, each action requires the agent to select an edge and compulsorily flip it. The termination time is the same as in Local and Global.

\begin{figure}[h]
    \centering

    \begin{subfigure}{0.35\textwidth}
        \centering
        \includegraphics[width=\textwidth]{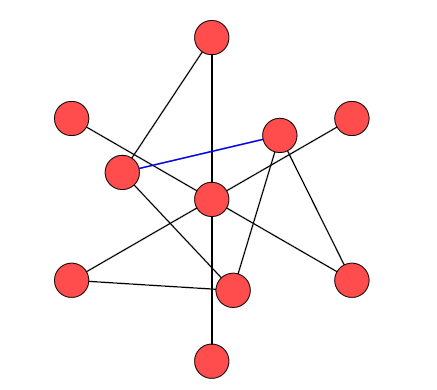}
        \caption{\textit{Linear}}
        \label{fig:subfigA}
    \end{subfigure}
    \hfill
    \begin{subfigure}{0.35\textwidth}
        \centering
        \includegraphics[width=\textwidth]{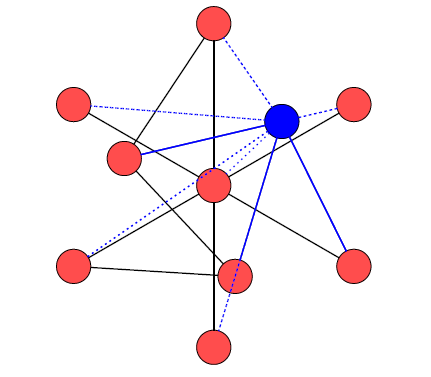}
        \caption{\textit{Local}}
        \label{fig:subfigB}
    \end{subfigure}
    \hfill
    \begin{subfigure}{0.33\textwidth}
        \centering
        \includegraphics[width=\textwidth]{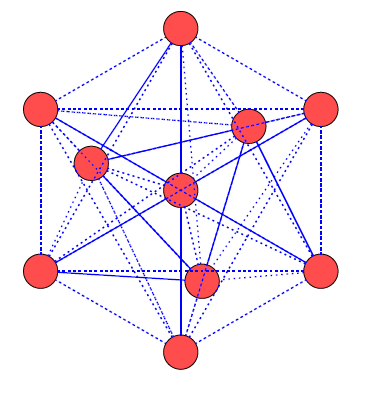}
        \caption{\textit{Global} and \textit{Flip}}
        \label{fig:subfigC}
    \end{subfigure}

    \caption{\textit{Comparison of game modes}. Edges that can be modified by the agent are colored blue, with dotted lines representing missing edges. In \textit{Linear} the agent can modify just the next edge in the game’s order. In \textit{Local}, the agent remains on vertices, such as the blue one in \ref{fig:subfigB}, and can choose to modify one among its incident edges. In \textit{Global} and \textit{Flip}, all edges are always accessible.}
    \label{fig:main}
\end{figure}

\subsection{Common settings}
\label{sec:common}

In this section we describe the parameters that can be used to tune specific aspects of the games.
The number of nodes and the reward function must be given in input, while the other ones, highlighted with $^*$, have default values that can be changed at game's initialization.
\begin{itemize}
    \item \textbf{Number of nodes:} Our games are thought to explore graphs with fixed number of vertices. It is not possible to enlarge or reduce this dimension while playing.
    \item \textbf{Reward Function:} Implementing environments parametric with respect to the reward makes their structure completely independent from the conjecture. Defined externally, the reward function takes an adjacency matrix and a boolean, outputting a numeric value. This boolean determines whether the reward should be normalized, a process which can affect neural network training. The environment has a normalize reward option, which is then internally passed to the reward function. Note that reward normalization depends on knowing the maximum possible reward, which isn't always feasible.
    \item \textbf{Normalize reward$^*$:} This boolean value is passed to the function used in computing rewards. Default value is False.
    \item \textbf{Reward type$^*$:} Incremental or sparse. Default is sparse.
    \item \textbf{Initial graph$^*$:} The game can be started from any graph. Default value is the complete graph.
    \item \textbf{Self-loops$^*$:} Self-loops are permitted in our environments, under the assumption that agents will avoid them when the scenario specifically requires graphs without self-loops. To enforce this rule and provide an additional layer of control, we have implemented a Boolean option. If set to False, the step method will not execute any action and will instead return an error message whenever an agent attempts to create a self-loop. The default value is False.
    \item \textbf{Check at every step$^*$:} Whether to check or not at every step for counterexamples. Default is False.
    \item \textbf{Termination time:} When episodes end. Default is $\frac{n(n-1)}{2}$ when self-loops is False, and $\frac{n(n+1)}{2}$ when self-loops is True.
\end{itemize}

\begin{remark}
Consider an agent playing a finite-time horizon ``build your graph'' game, with horizon $T$. An agent looking for an optimal graph $G^*$ maximizing a certain conjecture $f$ would need to synchronize its moves to conclude the episode exactly on $G^*$. If we assume for simplicity that the game starts from the empty graph, in scenarios where the only action is flipping an edge, without the option to pass, the parity of the number of edges of the final graph $G_T$ would be dictated by the parity of $T$. Thus, it may well be impossible, also for a perfect agent, to exploit $G^*$, without a pass action.
\end{remark}

\begin{remark}
We raise questions about the suitability of RL for this type of combinatorial optimization problem. The outcome of an RL algorithm is a policy, that can then be employed to play optimally. However, within this specific framework, the focus is not on the policy itself but rather on the final state that the policy produces.
We wonder whether a different approach, for instance a generative model, could be a more appropriate approach. Such a model would focus on iteratively constructing better graphs, which aligns more directly with the primary goal of discovering counterexamples or optimal structures without the intermediary step of policy refinement.
\end{remark}

\subsection{User interface}

To facilitate testing of the game environments described herein, we provide a simple graphical user interface (GUI), which is open-sourced and available at \url{https://github.com/CuriosAI/graph_conjectures} under \texttt{main.py}. This interface enables users to select a conjecture (Wagner or Brouwer), choose a game type (Linear, Local, Global, or Flip), specify the number of nodes, and select the reward type (Sparse or Incremental). Users can interact with the game by performing actions, with the GUI visualizing the current state of the game as graph $G$, as well as the function $f(G)$ that is intended to be maximized.

\section{Dataset for Laplacian spectrum supervised learning}
\label{sec:dataset}

In this section we describe a dataset built for working on  Brouwer's Conjecture~\cite{Brouw10}, see Section~\ref{sec:future}. The dataset contains graphs with $11$ nodes sampled from diverse distributions, and labelled with their Laplacian spectra. This number of nodes is the first one for which the conjecture has not yet been proved.

We used three different random graphs models implemented in NetworkX Python library and graphs downloaded from The House of Graphs database~\cite{HouseGr}. Our dataset contains:
\begin{itemize}
    \item 1010 graphs drawn from Erdős–Rényi (ER) models, denoted as $G(11, p)$. We considered $p$ varying in $[0,1]$ with step $0.01$, obtaining 99 non-trivial models, and two trivial distributions that create copies of the empty graph or the complete graph. We have drawn $10$ graphs from each of this models. Our choice to produce a small samples size at fixed $p$ is motivated by the fact that slightly similar probabilities $p_1$ and $p_2$ lead to similar ER models, and thus, to similar graphs. Small models' sample sizes produces a limited redundancy of configurations, resulting in a oversampling effect that helps the learning process.
    \item 540 graphs drawn from Watts-Strogatz models, varying the mean degree $k$ in $\{4,6,8\}$, avoiding $k=10$ to stay away from the complete graph, and rewriting edges' probability $\beta$ in $[0.1,0.9]$ with step $0.1$. Combining $k$ and $\beta$ in all possible ways gives 27 models and 20 graphs were drawn from each.
    \item 271 graphs with 11 nodes downloaded from The House of Graphs. House of Graphs is a rich database of non-isomorphic graphs which is frequently updated. It contains a lot of particular configuration that can be very hard to reach with random graph's generators.
    \item 10162 graphs obtained with Barabási–Albert model (BA), with parameter $m$ varying in $\{2,\dots,9\}$. BA algorithm builds a graph starting from an initial configuration on $m_0>m$ nodes. We took House of Graphs' samples with $3 \leq n \leq 10$ and used each $G$ in this batch to start a BA generation with $m<|V(G)|$.
\end{itemize}

The dataset is open-sourced and available at \url{https://github.com/CuriosAI/graph_conjectures} under the filename \texttt{n11\_graphs.g6}. The \texttt{.g6} format is a compact text-based encoding specifically designed for graphs. It is well-supported and can be easily read by the \texttt{NetworkX} method \texttt{read\_graph6}, that returns a list of graphs objects.
Labels are in \texttt{n11\_laplacian\_spectra.txt}. This is a simple text file, where each line includes 11 Laplacian eigenvalues in descending order, separated by spaces. Each line in the \texttt{n11\_laplacian\_spectra.txt} directly corresponds to the graph at the same position in the \texttt{n11\_graphs.g6} file.

All pairs of graphs in the dataset have been subjected to the 1-dimensional Weisfeiler-Leman test \cite{WeisLem}. This test gives a negative response in case of non-isomorphic graphs, and positive in case of potentially isomorphic graphs (false positives are possible). The test resulted positive on 1124695 pairs, meaning that the percentage of isomorphic pairs is less than or equal to the 1,57\% of the total. The highlighted pairs are reported in the file \texttt{weisfeiler\_leman\_results.txt}.

This dataset can be used in selecting GNNs for any conjecture regarding Laplacian eigenvalues, not only Brouwer's Conjecture.
In Section \ref{sec:future} we discuss how this dataset could be further improved.

\section{Future Work and Conclusions}
\label{sec:future}

An important next step for our proposed systematization is to integrate the computation of invariants using external Java code. As pointed out by~\cite{GAKRL1}, this could significantly enhance the computational efficiency.

An interesting direction for further research would be conducting an ablation study to evaluate the effectiveness of various components described in this paper. For instance, we could set a fixed conjecture, architecture, and algorithm, and then test Linear, Local, Global, and Flip, to see if one of them is particularly more effective than the others. This method could also be employed to assess the impact of other elements within the framework.

Additionally, we plan to apply AlphaZero \cite{SHSMCS} and other AlphaZero-like algorithms \cite{MAFSSAec,MAGSSA,PPMSWS} to the Brouwer conjecture, using our dataset to select an appropriate GNN flavor. We intend to experiment with the pairformer, implemented following the description in~\cite{MAKFIL} and the AlphaFold source code, from which the pairformer is taken. All neural network architectures will be enhanced by a global skip connection \cite{DMFGCN} and by non-shared biases \cite{MFPRCUwacv}.

Also the dataset could be improved. We plan to enhance its diversity by calculating a range of graph invariants to identify and downsample overrepresented elements. Additionally, we will explore new methods to enrich the dataset further.

We conclude with a very general remark. Wagner's approach holds potential beyond graph theory, and it could be applicable to any combinatorial optimization problem. In a sense, RL is a very powerful ``algorithm'' for discrete optimization, and Wagner's paper has ``just'' effectively exploited this power in the realm of graphs.


\bibliographystyle{splncs04}
\bibliography{flora.bib}

\end{document}